\documentclass{article}

    \PassOptionsToPackage{numbers, compress}{natbib}
\usepackage[preprint]{neurips_2024}

\usepackage[utf8]{inputenc} % allow utf-8 input
\usepackage[T1]{fontenc}    % use 8-bit T1 fonts
\usepackage{hyperref}       % hyperlinks
\usepackage{url}            % simple URL typesetting
\usepackage{booktabs}       % professional-quality tables
\usepackage{amsfonts}       % blackboard math symbols
\usepackage{nicefrac}       % compact symbols for 1/2, etc.
\usepackage{microtype}      % microtypography
\usepackage[pdftex]{graphicx}
\usepackage{xcolor}
\usepackage{multirow}
\usepackage{algorithm}
\usepackage[noend]{algpseudocode}
\usepackage{caption}
\usepackage{subcaption}
\usepackage{bm}
\usepackage{amsmath}
\usepackage{enumitem}
\usepackage{amsthm}
\DeclareMathOperator*{\argmax}{arg\,max}
\DeclareMathOperator*{\argmin}{arg\,min}

\title{Hybrid Learners Do Not Forget: A Brain-Inspired Neuro-Symbolic Approach to Continual Learning}

\author{%
  Amin Banayeeanzade \\
  Department of Computer Science\\
  University of Southern California\\
  \texttt{banayeea@usc.edu} \\
  \And
  Mohammad Rostami\\
  Department of Computer Science\\
  University of Southern California\\
  \texttt{rostamim@usu.edu} \\
}

\begin{document}

\maketitle

\begin{abstract}

Continual learning is crucial for creating AI agents that can learn and improve themselves autonomously. A primary challenge in continual learning is to learn new tasks without losing previously learned knowledge. Current continual learning methods primarily focus on enabling a neural network with mechanisms that mitigate forgetting effects. Inspired by the two distinct systems in the human brain, \textit{System 1} and \textit{System 2}, we propose a Neuro-Symbolic Brain-Inspired Continual Learning (NeSyBiCL) framework that incorporates two subsystems to solve continual learning: A neural network model responsible for quickly adapting to the most recent task, together with a symbolic reasoner responsible for retaining previously acquired knowledge from previous tasks. Moreover, we design an integration mechanism between these components to facilitate knowledge transfer from the symbolic reasoner to the neural network. We also introduce two compositional continual learning benchmarks and demonstrate that NeSyBiCL is effective and leads to superior performance compared to continual learning methods that merely rely on neural architectures to address forgetting.
\end{abstract}

\section{Introduction}

%Traditional machine learning methods assume that training and test data are drawn from the same distribution and that the model will not encounter new data outside this distribution. Continual learning is an effort to equip the learning algorithms with the ability to persistently learn new tasks in the environment while not catastrophically forgetting their previously acquired knowledge of the past tasks. In recent years, the community has devoted significant attention to mitigating the forgetting phenomena in neural networks. Despite their noteworthy accomplishments, we raise this fundamental research question: Are the systems purely made from neural networks the best candidates to endure the burden of continual learning? 

The classic approach for machine learning (ML) is to learn a single task in isolation. This assumption entails that the training and the test data are drawn from the same distribution. As a result, most ML models fail to generalize well when they encounter data outside the training data distribution. The naive solution to make a model continually generalizable is to continuously update the model to fit the input distribution. This process, however, will lead to performance degradation on past tasks, known as \textit{catastrophic forgetting} in the literature~\cite{french1999catastrophic}, because the learnable parameters are dragged away from optimal values for the past tasks. The goal of continual learning (CL) in deep learning is to mitigate forgetting effects when a model encounters sequentially arriving tasks~\cite{chen2018lifelong}. In recent years, the community has devoted significant attention to mitigating the forgetting phenomena in neural networks \cite{parisi2019survey}. Despite their noteworthy accomplishments, we raise this fundamental research question: \textit{Are the systems purely made from neural reasoners the best candidates to address CL? }

Humans, as natural intelligent agents, demonstrate exceptional cognitive abilities and have continuously inspired AI research. As hypothesized by \citet{kahneman2011thinking}, the human brain operates using two distinct systems, namely \textit{System 1} and \textit{System 2}. While System 1 is fast, automatic, and intuitive, System 2 is slower, deliberate, and analytical, employing logical reasoning to evaluate information and make conscious decisions. The two systems operate in parallel and are considered to be complementary problem solvers. During the initial stages of learning a task, System 2 plays a more prominent role. Learning a new skill requires deliberate practice, focus, and attention to the details, all of which are attributes of System 2. System 2 is responsible for consciously processing new information, breaking it into smaller meaningful parts, and forming new mental models and patterns. As the learning progresses, with practice and repetition, System 1 becomes more adept at performing the task automatically and without the help of System 2 \cite{EVANS2003454, BARROUILLET201179}. 

In the inference phase, System 1 thinks fast and is more likely to be engaged in tasks that are routine, habitual, and reflexive. 
System 1 works based on the formation of associations between stimuli and responses, which can become weaker over time if they are not reinforced through repeated exposure or use. In contrast, System 2 decision-making is based on the explicit, conscious processing of information, which can be beneficial when solving problems that require conscious effort, analytical thinking, and logical reasoning. Therefore, when facing out-of-distribution tasks, either novel or outdated, System 2 is more likely to be responsible for handling them \cite{kahneman2011thinking}. 

When comparing available paradigms in AI with the two systems in humans, artificial neural networks are believed to behave very similarly to System 1, while symbolic reasoners share similar characteristics with System 2. With this connection in mind, existing methods in the CL literature primarily focus on equipping System 1, i.e., neural networks, with the capability of addressing forgetting. In humans, however, this is not the case. While System 1 is vulnerable to forgetting, System 2 experiences minimum forgetting by leveraging a symbolic reasoning mechanism. As an example, consider the procedural learning of a motory skill, e.g., swimming. While during the early stages of learning, most of the muscle behaviors are derived deliberately by System 2, gradually, the skill consolidates into the unconscious structure of System 1, leading to generating the appropriate motory responses without explicit reasoning. However, if System 1 eventually forgets to solve the task due to discontinued exposure to performing the task over extended time periods, System 2 is still capable of bringing back the procedural sequence by which the learning has started before.
 
 To create adaptive AI systems, the emerging neuro-symbolic approaches~\cite{besold2017neural} integrate System 1, represented by a neural network, with System 2, implemented as a symbolic reasoner, to enable the AI system to reason about complex problems while learning from continuous experiences. Inspired by the recent success of neuro-symbolic ML~\cite{2021nssurvey, parisottoneuro, mao2019neuro}, we develop a Neuro-Symbolic Brain-Inspired Continual Learning (NeSyBiCL) framework to address CL by mimicking similar human-inspired mechanisms. Our framework is composed of three main subcomponents. First, a low-level feature extractor, implemented by a convolutional neural network (CNN), extracts basic patterns in the input. Second, a neural network architecture is used on top of the features to imitate the System 1 behavior by making quick and accurate inferences.
 Finally, a symbolic inference engine, namely a probabilistic symbolic reasoner, is used on top of the features to imitate System 2, which provides the agent with interpretable reasoning capability to mitigate catastrophic forgetting.
 
 The symbolic machinery is responsible for breaking the tasks into meaningful subconcepts and solving the tasks by conducting logical deductions. Since rule-based symbolic learners face serious limitations, a neural network learner is used to solve the most contemporary task at hand. While the neural network component excels at solving the most recent task, it is prone to forget the older ones. Although the symbolic reasoner is not as accurate as the neural network, we demonstrate that it is immune to catastrophic forgetting. In addition to this property, we propose an interaction mechanism between the two systems, enabling an effective knowledge transfer between both learners. 
 
Notwithstanding the desired properties, our model is only applicable to tasks with a compositional nature, i.e., the tasks that can be solved by decomposing the input into smaller subconcepts and drawing conclusions based on them and their relationships. To provide further insights, we introduce two novel benchmarks to demonstrate the effectiveness of NeSyBiCL compared to existing neural-based CL methods. 

\section{Related Work}

Our work is at the intersection of continual learning and neuro-symbolic AI. We review relevant works in these areas as well as compositional tasks in AI, which are great use cases for our algorithm. 

\paragraph{Continual Learning} Continual learning, also known as continual learning,   addresses the challenge of acquiring new knowledge by a model while retaining previously learned knowledge. To tackle the issue of forgetting,  three main approaches have been proposed \citep{parisi2019survey}.
The first approach, known as replay-based methods, involves reusing samples from previous training tasks. This can be achieved by either storing them in an explicit episodic memory \citep{Gem,icarl,bilevel,gradient_based} or generating pseudo-samples using a deep generative model \citep{deep_gen,MREGan,brainrep}. Restored samples can then be mixed with new data and replayed during model update episodes, simulating joint training \citep{MREGan, deep_gen}, or used as a constraint to regularize the update of model parameters \citep{Gem}.
The second approach uses parameter isolation which focuses on dynamically expanding the model as new task data arrives. This expansion involves introducing task-specific weights \citep{pnn, DEN, progress} or masks \citep{supsup, order_rob} to accommodate the new tasks effectively.
The third approach is based on using regularization on the model which involves penalizing significant deviations of parameters from the optimal solutions for the previous tasks \citep{VCL, ZenkePG17, kirkpatrick2017overcoming}. 

In this context, the proposed method presented in the paper differs substantially from the aforementioned approaches. While the existing methods concentrate on equipping neural networks with mechanisms to prevent catastrophic forgetting, we argue that neural networks are not the best candidates for continual learning. Instead, we introduce a symbolic mechanism as the solution to continual learning challenges. By adopting this symbolic approach, we offer an alternative perspective on addressing continual learning.

\paragraph{Neuro-Symbolic AI} The goal of neuro-symbolic AI is to combine neural networks and symbolic reasoning techniques to learn complex tasks. This interdisciplinary approach has shown promising results in various domains, including question answering \citep{yi2018neural}, natural language understanding \citep{liu2022NSNLU}, and logical reasoning \citep{manhaeve2018deepproblog}.
In the context of continual learning, there have been preliminary efforts to leverage symbolic representations. One example is the use of symbolic machinery to remember previous tasks in reinforcement learning domains \citep{silver2013common}. This approach demonstrates the effectiveness of symbolic representations in retaining knowledge across different reinforcement learning tasks.
Another approach is to combine symbolic reasoning with the representational capabilities of neural networks to develop continual reasoning agents \citep{marconato2023neuro}. By integrating symbolic reasoning and neural networks, the system achieves a more comprehensive and flexible approach to continual learning. These methods do not rely on their symbolic reasoner for the purpose of combatting catastrophic forgetting, but rather, they rely on the existing CL strategies, such as rehearsal, as the main strategy against forgetting. In other words, the reason behind using symbolic reasoners in these studies is only because they fit logical reasoning tasks better. 

In contrast, our work specifically draws inspiration from the concept of ``slow and fast thinking'' theory of the brain \citep{bengio2019consciousness, chen2019deep}. This theory suggests that the brain employs two distinct processes for data processing: a fast, intuitive process and a slower, deliberate process \cite{kahneman2011thinking}. In the context of continual learning, we aim to leverage symbolic reasoning to mitigate catastrophic forgetting.
By capitalizing on the benefits of both symbolic reasoning and neural networks, we develop a framework that takes advantage of the dual process theory in the brain.

\paragraph{Visual Compositionality}

Visual compositionality refers to the ability to understand and interpret visual scenes by decomposing them into meaningful components or objects and analyzing their relationships and interactions. Compositionality is believed to be the key to the next level of intelligence and has attracted many research endeavors. For example, previous works have used EBMs for compositional image generation \cite{Li2020compositional}, zero-shot image classification \cite{wu2022zeroc, Han2023} and visual question answering \cite{Andreas2016NMN}. Compositionality has also been studied in continual learning by introducing compositional modular structures \cite{mendez2021lifelong}, or by relying on EBMs to enable continual learning \cite{Li2020compositional}. Our work also follows the same goal of using compositionality in continual learning but with a focus on symbolic methods rather than pure neural networks.

\section{Preliminaries}

Consider a sequence of $t = 1,\cdots,T$ learning tasks with the training datasets $\mathcal{D}_t = \{(\bm{x}_i,\bm{y}_i)\}^{i=1}_N$, where $\bm{x}_i \in \mathcal{X}$ is the input sample, and $\bm{y}_i \in \mathcal{Y}$ is its corresponding label, drawn from an unknown task-dependent distribution as $(\bm{x},\bm{y}) \sim p^{(t)}(X,Y)$ 
\footnote{With little misuse of notation, we may use $\bm{y} \in \mathcal{D}_{\leq t}$ to show that the label of class $\bm{y}$ exists in the tasks from the beginning up to the timestep $t$.}.
At each timestep, the agent learns the current task and then proceeds to learn the next task. 

\begin{figure} 
  \centering
  \includegraphics[width=0.99\linewidth]{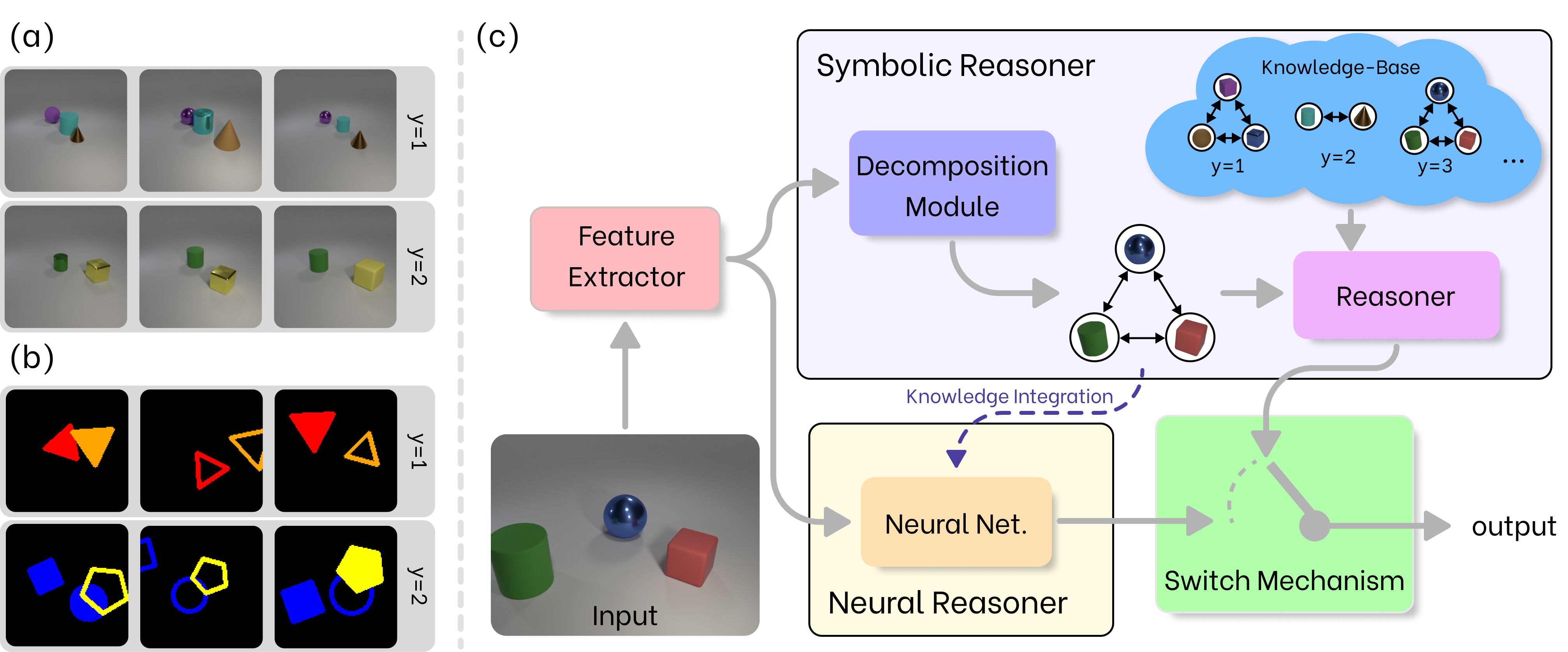}
   \caption{ An overview of the proposed framework.  (a)-(b)  examples from the Clevr and Sketch compositional datasets. In both datasets, the combination of objects (subconcepts) with different shapes, colors, and relational locations form a new class, while other factors, such as rotation, color fullness, or material, are irrelevant. As an example, in the Sketch domain, a red triangle to the right of an orange one forms one class, while the combination of a blue circle, a yellow pentagon on the right, and a blue rectangle on the left creates another class. (c) A schematic of NeSyBiCL which consists of two complementary parallel pathways for data processing. 
   Inspired by the dual thinking system, the symbolic reasoner decomposes the input into its subcomponents/relations and is the best candidate to solve the older tasks with no forgetting. The neural reasoner is responsible for effectively solving the contemporary task. The two components are also coupled through a knowledge integration loss.
   }
   \vspace{-3mm}
\label{fig:method}  
\end{figure}

In this work, we consider tasks that have a compositional nature, i.e., the tasks could be 
solved more effectively by breaking them down into smaller parts and then combining the partial solutions through their relations to obtain the final solution. For example, see Figure~\ref{fig:method}~(a)-(b), where each class is defined by the existence of a number of parts (abstract subconcepts) in the input. The smaller parts in the task have bilateral relations that make reasoning about solving the tasks plausible. 
In other words, each class in the dataset can be characterized by a graph that encodes the compositional knowledge. As a result, one can reason on the set of graphs to distinguish the classes from each other. 
Leveraging compositionality and abstract subconcepts for relational reasoning has been shown to significantly enhance generalization in AI ~\cite{ito2022compositional, wu2022zeroc}. However, this mechanism has not been widely adopted in CL settings \cite{Li2020compositional}. 

The goal in CL is to sequentially observe the tasks and train a predictive model $\hat{\bm{y}} = h_\theta(\bm{x};t)$ with learnable weights $\theta$, such that it generalizes well on all tasks observed so far continually during the testing time. 
Since the agent may encounter the past learned task at any timestep, it should preserve its generalizability on those tasks. Simply updating the model at each timestep using empirical risk minimization on $\mathcal{D}_t$ will lead to catastrophic forgetting. Because the model will be trained to perform well only on the current task, leading to the deviation of $\theta$ from the optimal values for past tasks. 

Existing CL methods mostly use neural-level approaches, i.e., equipping System 1 with effective mechanisms, such as model regularization or experience replay to mitigate forgetting~\cite{kirkpatrick2017overcoming,rolnick2019experience,aljundi2018memory, parisi2019survey}. However, our goal is to implement System 2 and take advantage of it to address catastrophic forgetting. At the same time, we also use System 1 to handle the uncertainty that the symbolic system is not able to capture due to the randomness of the tasks. This randomness may include limited spatial translations, occlusions, rotations, or choice of materials. System 1 has a more generalizable behavior to handle these uncertainties and can complement System 2. To further clarify the motivation behind this setup, consider the cognitive task of distinguishing an elephant from an eagle as an example. Humans can solve the task at a glance by intuitively using the System 1 prediction. However, one may discriminate between the two categories by probing the lack or existence of subconcepts, e.g., trunk, long ears, beak, etc, and their bilateral relationships. Therefore, the ability to think compositionally relies on the operation of System 2~\cite{Han2023}. 

\section{Proposed Method}

%In this section, we present the proposed symbolic reasoner, followed by a subsection on the neural reasoner. Next, we describe the training and inference procedure of the two systems.

Figure~\ref{fig:method}~(c) visualizes a block diagram representation of the proposed solution for CL, where a symbolic reasoner is used along with a neural reasoner. The first component in the computational pipeline is a feature extractor $f:\mathcal{X} \to \mathbb{R}^d$, implemented by a CNN, to embed the input in a $d$-dimensional feature space. Moreover, neuroscience studies have reported that the primary visual cortex of humans appears to remain unaltered beyond a certain age \cite{shaw2008neurodevelopmental,huttenlocher1979synaptic}, indicating that fundamental image feature extractors in the brain undergo infrequent rewiring. We have reflected this property by pretraining a feature extractor and fixing it as a basic processing block. On top of that, we employ two subsystems, i.e., the neural reasoner and the symbolic reasoner, as further explained in the next sections.

\subsection{Symbolic Reasoner} \label{sec:symb}

The key aspect of the symbolic reasoner is to establish a connection between a raw input $\bm{x}$ and its label $\bm{y}$, by inferring an intermediate symbolic abstract representation. We consider that this representation is encoded via a graph $\mathcal{G} = (\mathcal{C},\mathcal{R})$. Formally, the nodes $\mathcal{C}$ in the graph represent the subconcepts in $\bm{x}$, and the edges $\mathcal{R}$ depict the relations between the nodes, e.g., the spatial relative structure among the subconcepts. In other words, each class in the problem is defined by a specific graph. Hence, the symbolic reasoner uses the extracted graph to predict the class label via graph matching, i.e., symbolic reasoning. With this formalism in mind, we assume the following joint distribution over variables:

\begin{equation}
\label{eq:factorization}
    p(\bm{x}, \mathcal{G}, \bm{y} ; \text{KB}^{(t)}) = p(\bm{y} | \mathcal{G} ; \text{KB}^{(t)}) p(\mathcal{G} | \bm{x}) p(\bm{x})
\end{equation}

where $\text{KB}^{(t)}$ is the abstract knowledge-base, accumulated up to timestep $t$ during the continual learning procedure of the symbolic reasoner. One implementation for such knowledge-base $\text{KB}^{(t)}$ could be the set of the learned class-specific graphs and their corresponding ground truth labels, i.e. $\text{KB}^{(t)} = \{(\bm{y},\mathcal{G}_y) | \bm{y} \in \mathcal{D}_{\leq t}\}$. The learning procedure for the symbolic reasoner is further explained in Section \ref{sec:learning}, but in a nutshell, the symbolic reasoner learns how to abstract a class-specific graph that entails the similarities between all class members in the training data when learning the current task. 
With the factorization in Equation~\eqref{eq:factorization}, the classification rule for symbolic reasoning would be:

\begin{equation}
\label{eq:classification}
    p(\bm{y} | \bm{x} ; \text{KB}^{(t)}) = \sum_{\mathcal{G} \in \mathbb{G}} p(\bm{y} | \mathcal{G} ; \text{KB}^{(t)}) p(\mathcal{G} | \bm{x}),
\end{equation}
where $\mathbb{G}$ is the set of all possible graph formations. Next, we estimate $p(\bm{y} | \mathcal{G} ; \text{KB}^{(t)})$ as $\text{softmax}(\text{sim}(\mathcal{G},\mathcal{G}_y))$, where $\textit{sim}$ is a similarity score function defined over $\mathbb{G} \times \mathbb{G}$. Moreover, one might assume the MAP approximation for simplification of solving Equation~\eqref{eq:classification}:

\begin{equation}
\label{eq:mapclassification}
    p(\bm{y} | \bm{x} ; \text{KB}^{(t)}) \approx p(\bm{y} | \mathcal{G}^*(\bm{x}) ; \text{KB}^{(t)}),  \text{ where }  \mathcal{G}^*(\bm{x}) = \argmax_{\mathcal{G}} p(\mathcal{G} | \bm{x})
\end{equation}

To model the graph construction from the input, we use the $p(\mathcal{G}|\bm {x}) = \prod_{\bm{c}\in \mathcal{C}} p(\bm{c}|\bm{x}) \prod_{\bm{r}\in \mathcal{R}} p(\bm{r}|\bm{x})$ factorization, and implement a decomposition module $D:\mathbb{R}^d \to \mathbb{G}$, e.g., a Faster R-CNN object detector, to decompose the input image into its subcomponents. The subconcepts that are abstracted from the inputs and the type of bilateral relations may be shared across various tasks.  They are crucial for exchanging knowledge across different classes, distinguishing them from each other, and enabling high-level reasoning \cite{gruber1995toward}. They are analogous to the symbols that humans use as a lingua-franca for the reuse of knowledge~\cite{kambhampati2022symbols, marconato2023neuro}. We adopt the inherent assumption that the meanings of concepts (symbols) for different tasks remain unchanged over time, making the symbolic reasoner temporally robust. In other words, we assume that the probability distribution $p(\mathcal{G}|\bm {x})$, have no time-varying element. With these assumptions, we come to the central claim of our work.

\newtheorem{statement}{Statement}
\begin{statement}\label{statement}
Under the assumption of a persistent decomposition module, the continual learner given in Equations~\eqref{eq:classification}~or~\eqref{eq:mapclassification} has zero forgetting.
\end{statement}

The above statement, with the proof included in Appendix~\ref{sec:proof}, justifies the intuition behind selecting the symbolic reasoner as the candidate to accumulate knowledge over a long continual episode. Notice that the persistency assumption is not very restricting since it has been reported that while the final semantic layers of neural networks drift considerably during a continual learning episode, the initial layers remain unchanged~\cite{lesort2021continual, ramasesh2021anatomy}. This phenomenon even enables the possibility of replacing the initial layers with a pretrained fixed feature extractor \cite{gemcl2021, ostapenko2022continual}. Put differently, in this work, we substituted the MLP layers on top of feature extractors with a symbolic reasoner to mitigate the problem of catastrophic forgetting. 

\subsection{Neural Reasoner}

The neural reasoner is a connectionist model~\cite{mcclelland1995there}, implemented with MLP layers on top of the feature extractor layers. It is responsible for directly mapping the features into the target space and making intuitive predictions using low-level features. More concisely, considering an embedding space of size $p$, we denote the neural network parameterized by $\theta_n$ as $h_{\theta_n}: \mathbb{R}^d \to \mathbb{R}^p$ followed by a logistic regression layer with the weights $W \in \mathbb{R}^{|\mathcal{Y}|\times p}$ and the bias term $b \in \mathbb{R}^{|\mathcal{Y}|}$ to generate an output label distribution as $\text{softmax} (Wh_{\theta_n}(\bm{x}) + b)$.  Existing AI systems primarily only have this component which is vulnerable with respect to catastrophic forgetting in CL settings. For this reason, existing CL methods focus on mitigating forgetting in this component by changing the end-to-end learning process of deep learning.
% Inspired by the interaction between System 1 and System 2, we propose an integration mechanism to facilitate knowledge integration from the symbolic reasoner into the neural network to tackle catastrophic forgetting using symbolic reasoning. Our core idea is to enforce the neural reasoner to encode the abstract information available by the symbolic reasoner. 

\subsection{Learning Process} \label{sec:learning}

\paragraph{Symbolic Reasoner:} At the timestep $t$ of the training phase, all of the training samples in $\mathcal{D}_t$ are parsed into their corresponding relational graphs. This process requires a priori information about the type of components and relations which describe the tasks well.  Note, however, the natural learning process relies on a similar premise. Next, the extracted graphs are analyzed and the graph with the minimum distance to other instances is assumed as the prototype graph that characterizes that class. The new graph then is appended to the knowledge-base (refer to Algorithm \ref{alg:learning} for more details).

\paragraph{Neural Reasoner:} We would like to augment the classic end-to-end learning process for deep learning by an interaction process between System 1 and System 2. Therefore,  we propose an integration loss $\mathcal{L}_\mathcal{I}$ to facilitate knowledge integration from the symbolic reasoner into the neural network. To that end, we utilize an MLP head parameterized by $\theta_i$ and apply it to the embedding space of the neural reasoner to predict some summarized attributes of the graph $\mathcal{G}$. As a result, the embedding space would encode the abstract information available in the input. Intuitively, this loss function encourages the neural learner to benefit from the human-understandable supervision provided by the symbolic reasoner. We also use the typical cross-entropy loss, $\mathcal{L}_\mathcal{C}$. The total loss function is a weighted combination of the two loss functions
\begin{equation} \label{eq:loss}
\mathcal{L}(\theta_n, \theta_i, W, b) = \mathcal{L}_\mathcal{C}(\theta_n, W, b) + \lambda \mathcal{L}_\mathcal{I}(\theta_n, \theta_i),
\end{equation}

where $\lambda$ is a trade-off hyperparameter between the two loss terms (see Appendix~\ref{sec:int_loss} for extra details). 

The neural reasoner is merely trained on the samples of the most recent tasks and is not responsible for remembering the older tasks. Therefore, we empirically found it beneficial to reset the $\theta_n$ and $W$ parameters at the beginning of each task to increase the learning capacity of the neural network. Notice that this technique only applies to classification heads and not the feature extractor to maintain the representational knowledge transfer between the tasks. Algorithm \ref{alg:learning} summarizes the learning process for NeSyBiCL.

\begin{algorithm}
\caption{Learning Phase of NeSyBiCL}\label{alg:learning}
\begin{algorithmic}[1]
\Require A pretrain dataset $\mathcal{D}_0$ and a set of $T$ continual tasks $\mathcal{D}_1, ..., \mathcal{D}_T$

\State $\text{KB}^{(0)} = \{\}$
\State Pretrain a feature extractor $f$ and a decomposition module $D$ with $\mathcal{D}_0$
\For {$t = 1$ to $T$}:
\For {$y \in \mathcal{D}_t$}:
\State $G = \{\mathcal{G}^*(x) | (x,y) \in \mathcal{D}_t \}$
\State $\mathcal{G}_y = \argmin_{\mathcal{G} \in G}
\sum_{\mathcal{G}' \in G} \text{sim}(\mathcal{G},\mathcal{G}')$
\EndFor
\State $\text{KB}^{(t)} \leftarrow \text{KB}^{(t-1)} \cup \{(y, \mathcal{G}_y) | y \in \mathcal{D}_t\}$ 
\State Reset $W, b, \theta_n$ and $\theta_i$
\State Optimize the neural reasoner using the loss function in Equation~\eqref{eq:loss}

\EndFor

\State \Return $f$, $D$, $\text{KB}^{(T)}$, $\theta_n$, $W$ and $b$
\end{algorithmic}
\end{algorithm}

\subsection{Inference}

Contemporary and old tasks are distributed differently among System 1 and System 2. System 1 is responsible for processing habitual tasks, which are familiar and require little conscious effort. In contrast, System 2 is used to process old tasks, which are less familiar and require deliberate thinking and effort. Based on this inspiration, we employ a switching mechanism to assign different tasks to different systems at the test time, as proposed in Algorithm~\ref{alg:inference}. Namely, the neural reasoner is the most effective module to handle the most recent tasks, while the symbolic reasoner is the best candidate to solve the older tasks without exhibiting catastrophic forgetting.

\begin{algorithm}[!h]
\caption{Inference Phase of NeSyBiCL}\label{alg:inference}
\begin{algorithmic}[1]
\Require An input $\bm{x}$ and the task id $t$
\Require The knowledge-base $\text{KB}^{(T)}$, the feature extractor $f$, decomposition module $D$ and the trained weights of the neural reasoner $W, b,$ and $\theta_n$.

\If {t = $T$}: 
\Comment Solving the most recent task
\State  $\bm{y} = \argmax  \; \text{softmax} (W h_{\theta_n} (f(\bm{x})) + b)$
\Else
\Comment Solving older tasks
\State $\mathcal{G}^*(\bm{x}) = D(f(\bm{x}))$
\State $\bm{y} = \argmax_{y, (y, \mathcal{G}_y) \in \text{KB}^{(T)}}  \text{softmax} (\text{sim}(\mathcal{G}^*(x), \mathcal{G}_y))$
\EndIf
\Return $\bm{y}$
\end{algorithmic}
\end{algorithm}

\section{Experimental Validation}

We demonstrate that  NeSyBiC leads to SOTA performance on two compositional CL benchmarks that we introduce. 

\subsection{Experimental Setup}

%In this section, we perform several experiments around our neuro-symbolic model to answer the following questions: (a) Can the hybrid model effectively solve learning tasks in a continual learning problem? (b) Is the symbolic learner immune to catastrophic forgetting, especially in long continual episodes? (c) Are both the neural reasoner and the symbolic reasoner essential for task-solving? (d) Under which circumstances is the integration loss helpful? To empirically answer the questions, 

\textbf{Datasets:} We designed two datasets, both with a compositional nature, using the \textit{Clevr} and \textit{Sketch} domains. The Clevr \cite {johnson2017clevr} dataset is a common dataset with a compositional nature usually used for visual question-answering tasks. Here, we synthetically arranged objects in each image to construct a continual dataset. We have also created a second dataset, called Sketch, to create similar images in a 2D scene. The shared property of these datasets is that each input image consists of several objects that possess certain bilateral relationships. Each class is defined by a class-level graph that is shared among the samples of that class. However, there exists a limited amount of variations in each image, e.g. translational randomness in the location of objects, which deforms the graph and makes it more challenging to reason about the class label. Figure \ref{fig:method} provides illustrations for samples of input images. We performed experiments using 10 continual tasks, with each task containing 10 classes. Each class was supported by 200 training samples and 50 test samples (see Appendix~\ref{sec:exp_details} for more details).

\textbf{Neural network architectures:}  We use a fixed CNN as the feature extractor network, followed by  a two-layers MLP for the neural reasoner which receives its input from the CNN. 
%We optimized neural architectures using the Adam optimizer and a learning rate of $10^{-3}$ for 200 epochs per task. It is noteworthy that each experiment is repeated 4 times and its average and variation is reported. (see supplementary for more details).

\textbf{Evaluation metrics:} let $\mathcal{R}_{i,j}$ denotes the model accuracy (classification rate)  on task $i$ after learning task $j$ in a CL setting. Following the literature, we use two metrics for evaluating the algorithms. First, we use the average accuracy over all learned tasks evaluated at the end of the continual episode when all tasks are learned, i.e. $\mathcal{A}_{\text{all}} = \frac{1}{T} \sum_{i=1}^{T} \mathcal{R}_{i,T}$. This metric is used to measure forgetting because it entails performance on all past tasks. Second, we also calculate the average of all accuracies on the most recent tasks immediately after the task is processed, i.e. $\mathcal{A}_{\text{last}} = \frac{1}{T} \sum_{i=1}^{T} \mathcal{R}_{i,i}$. This metric is a reflection of the model's adaptability and can be used to reveal the weakness of the symbolic learner. We repeat each experiment 4 times and report both the average and the standard deviation for these metrics. To study the temporal dynamics of learning, we also visualize learning curves.

\textbf{Baselines for comparison:}  
to demonstrate the effectiveness of our algorithm, we include a naive sequential learner, i.e., simple fine-tuning, as a lower-bound as well as a multi-task learner, i.e., joint-task training, as an upper-bound for CL algorithms.  Despite being simple, these baselines help to evaluate the effectiveness of our algorithm. Moreover, we include the results of our hybrid model, NeSyBiCL, in addition to a pure symbolic baseline, which is equivalent to removing the neural reasoner from the hybrid model and solely relying on the symbolic mechanism to make decisions for all tasks. This baseline can demonstrate the positive effect of using a neural reasoner.

For comparison against existing CL methods,  we compare our method against: (i) Experience Replay (ER)~\cite{rolnick2019experience}, (ii) Gradient Episodic Memory (GEM)~\cite{lopez2017gradient}, (iii) Elastic Weight Consolidation (EWC)~\cite{kirkpatrick2017overcoming}, (iv) Synaptic Intelligence (SI) \cite{ZenkePG17}, and (v) Learning without Forgetting (LWF) \cite{LiH16eLWF}. These methods are served as representatives of the major existing neural-based CL approaches. ER randomly stores a small subset of  training samples in a memory buffer, replayed when the model is updated for rehearsal. GEM uses memory samples for gradient correction when updating the model weights for the new task. EWC regularizes the loss function by penalizing the drift of network parameters from the weights optimal for previous tasks. SI adopts a similar regularization mechanism, except that it estimates parameter importance online based on 
the training trajectory. LWF, uses the inputs from the current task as anchor 
points and minimze the distance between the predicted logit vectors of the new model and the model from the previous timestep.

These techniques have their own limitations. For example, ER will not work well if a memory buffer with a fixed size is used or EWC compromises the learning capacity of the neural reasoner as more weights are consolidated to remember past learned tasks. In contrast,  NeSyBiC  stores an abstract knowledge graph in its symbolic memory as a lightweight prototype for each class, rather than input images in the pixel space. For a fair comparison, we conduct all  experiments on s single-head neural networks \footnote{except for the multi-task learning baseline where the backbone is shared and each task has a separate classifier head} in the task-incremental CL scenario~\cite{van2019three}.

\subsection{Results}

 Table~\ref{tab:result} presents a comparison between NeSyBiC and the baselines on both datasets. We observe that the symbolic reasoner excels other baselines in the average accuracy measure $\mathcal{A}_{\text{all}}$ by a significant performance gap. The underlying reason behind this behavior is that the symbolic learner can effectively break the input into meaningful high-level subconcepts and solve the tasks by exploiting 
 the compositional nature of the problem. However, due to the uncertainty in the tasks,  the predicted graph is different from the true graph for a subset of input samples. As a result, the symbolic reasoner is incapable of solving the contemporary task with a performance similar to the neural-based CL methods as reflected by the $\mathcal{A}_{\text{last}}$ measure. This trend is the primary reason that most symbolic learning is replaced with deep learning. In other words, in a single-task learning scenario, deep learning is a better solution. These observations demonstrate that AI exclusively based on System 1 or System 2 leads to specific weaknesses in CL settings. Importantly, we observe that NeSyBiC, which benefits from both System 1 and System 2, leads to superior performance in both measures. We conclude that while neural-based CL techniques are helpful, neuro-symbolic CL is an alternative with great potential.

\begin{table}[!t] 
  \centering
\begin{tabular}{@{}ccccc@{}}
\specialrule{1pt}{2.1pt}{2.5pt}
\phantom{Methods} & \multicolumn{2}{c}{Sketch} & \multicolumn{2}{c}{Clevr}        \\ \cmidrule(r){2-3} \cmidrule(r){4-5} 
Methods                  & $\mathcal{A}_{all}$(\%)   & $\mathcal{A}_{last}$(\%)   & $\mathcal{A}_{all}$(\%)   & $\mathcal{A}_{last}$(\%) \\ \midrule
Fine-Tuning                    & $18.2 \pm 0.5$           & $89.1 \pm 5.1$            & $19.2 \pm 1.0$              & $92.2 \pm 2.6$               \\
Multi-Task                    & $92.2 \pm 0.2$           & $-$                       & $94.5 \pm 0.3$              & $-$                          \\\cmidrule(r){1-5} 
EWC \cite{kirkpatrick2017overcoming}                      & $28.0 \pm 0.7$           & $\mathbf{90.8 \pm 1.6}$   & $27.1 \pm 0.3$              & $90.8 \pm 2.0$               \\
ER \cite{rolnick2019experience}                       & $35.0 \pm 1.3$           & $83.4 \pm 7.5$            & $31.4 \pm 3.2$              & $83.2 \pm 8.8$               \\
SI \cite{huttenlocher1979synaptic}                       & $23.6 \pm 1.0$           & $88.3 \pm 5.1$            & $25.9 \pm 0.7$              & $88.0 \pm 1.7$               \\
LWF \cite{LiH16eLWF}                      & $29.1 \pm 0.3$           & $88.3 \pm 5.1$            & $29.0 \pm 0.2$              & $87.1 \pm 5.8$               \\
GEM \cite{lopez2017gradient}                      & $22.4 \pm 1.1$           & $89.5 \pm 5.5$            & $25.1 \pm 4.6$              & $\mathbf{92.3 \pm 2.8}$               \\ \cmidrule(r){1-5}
Symbolic         & $67.2 \pm 0.5$           & $67.2 \pm 5.9$            & $64.6 \pm 1.6$              & $64.6 \pm 5.5$               \\
\hline
NeSyBiC                  & $\mathbf{71.5 \pm 0.6}$           & $90.6 \pm 5.0$            & $\mathbf{68.4 \pm 0.9}$              & $90.8 \pm 4.0$               \\ \bottomrule
\end{tabular} \vspace{0.9em}
\caption{Performance comparison of NeSyBiC along with baselines on the two compositional datasets:  symbolic reasoner faces no forgetting effects and   outperforms the neural-based CL baseline in the average accuracy criteria  $\mathcal{A}_{all}$. NeSyBiC hybrid-based learner utilize a neural reasoner parallel to the symbolic reasoner which enhances the adaptability to a new task, measured by the $\mathcal{A}_{\text{last}}$ criteria.  }
  \label{tab:result}
  \vspace{-4mm}
\end{table}

%$\mathcal{A}_{all}$ measures the average accuracy of the model among all continual tasks at the end of the episode and $\mathcal{A}_{last}$ evaluate the adaptability of the model to a new task by recording the model's performance immediately after the task is processed. Symbolic reasoner has no forgetting and therefore it outperforms other baselines with a great gap in the average accuracy criteria. Hybrid learner, utilize a neural reasoner parallel to the symbolic reasoner to enhance the adaptability to a new task

\begin{figure}[t]
\centering
	\begin{subfigure}{.44\textwidth}
		\centering
		\includegraphics[width=0.99\linewidth]{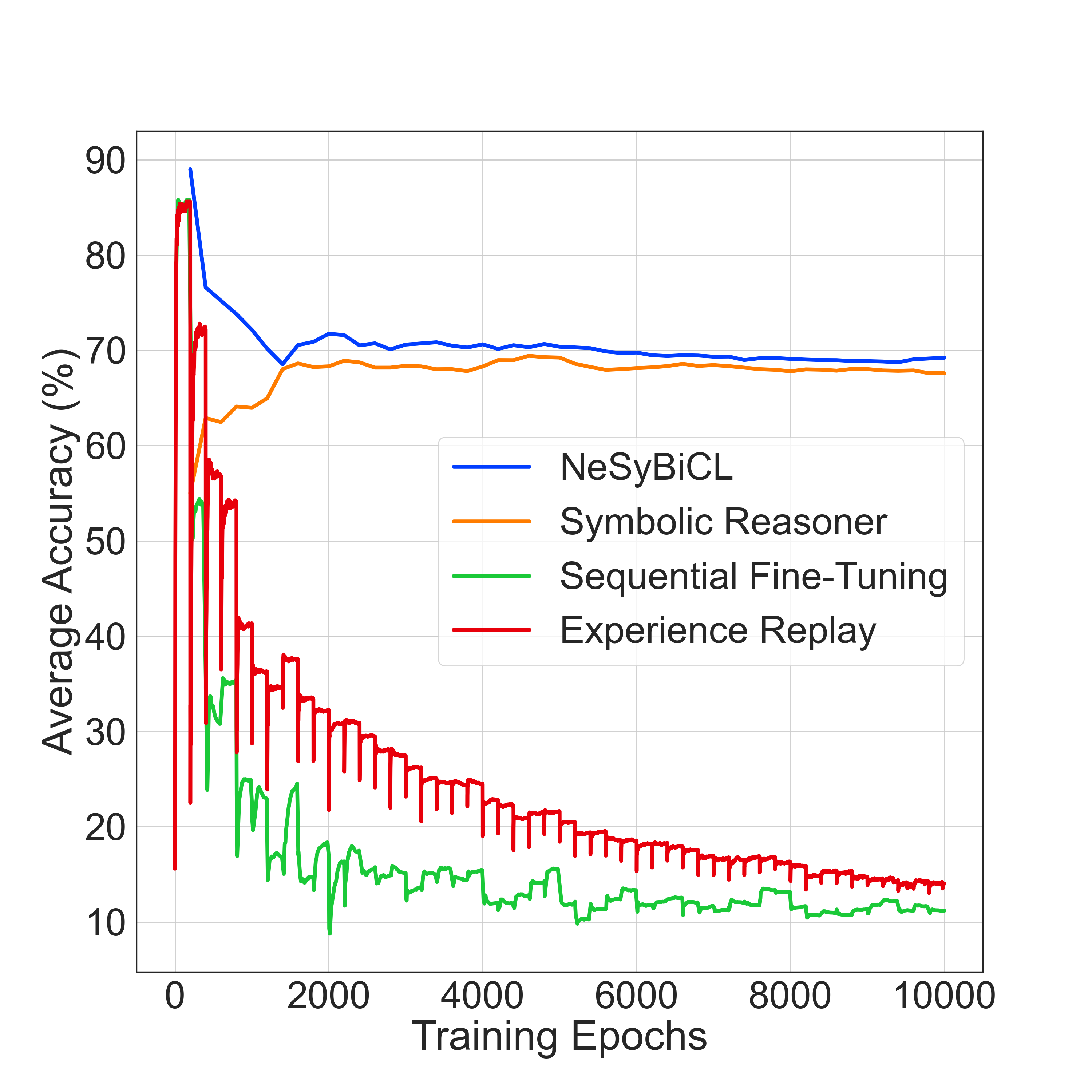}  
		\vspace{-0.35\baselineskip}
	\end{subfigure}
	\begin{subfigure}{.44\textwidth}
		\centering
		\includegraphics[width=0.99\linewidth]{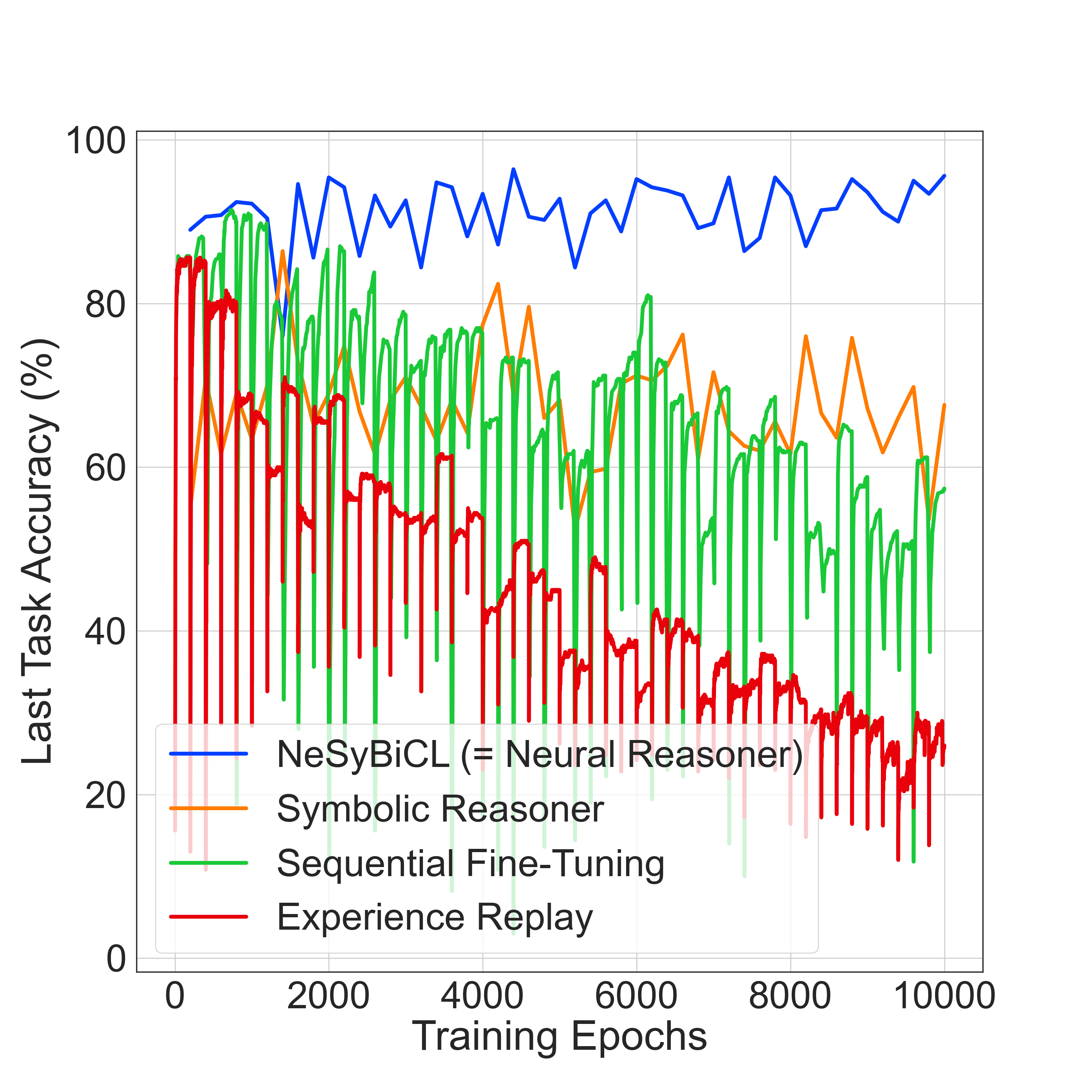}  
		\vspace{-0.35\baselineskip}
	\end{subfigure}
	\caption{
	Learning curves for a long continual learning episode consisting of 50 tasks in the Sketch domain:  \textbf{(Left)} the average accuracy on all previous tasks at each time step. We observe that NeSyBiC and the symbolic reasoner lead to minimal forgetting effects. \textbf{(Right)} the accuracy on the test samples of the last task at each timestep. We observe that NeSyBiC achieves better performance when solving the contemporary task compared to the symbolic learner.
	}
	\label{fig:episode}
    \vspace{-3mm}
\end{figure}

A major evaluation criterion in CL is the amount of forgetting. To study forgetting, we use a long continual episode based on the Sketch dataset. It consists of 50 tasks with 10 classes per task. To study the dynamics of learning, we have visualized the learning curves for this experiment in Figure \ref{fig:episode}. In this figure, neural-based CL methods are trained for 200 epochs on each task before the next task arrives.  Figure \ref{fig:episode}, left, presents $\mathcal{A}_{\text{all}}$ metric versus training epochs. In this figure, the near-flat behavior in the learning curve of the symbolic reasoner stems from its zero-forgetting property according to Statement~\ref{statement}. More importantly, we observe that although neural-based CL methods can slow down the forgetting effect when compared to the fine-tuning baseline, they are not as effective when used in a long CL episode. Particularly, we observe that at the end of the learning episode, the performance of ER is similar to fine-tuning. As expected, NeSyBiC outperforms both the symbolic and the neural reasoner techniques by combining their nice behaviors.

We provide the learning curves for the $\mathcal{A}_{\text{last}}$ metric in Figure \ref{fig:episode}, right, to study the plasticity-stability tradeoff in the CL methods. We observe that despite using a growing memory buffer for ER, the performance on the last task decreases. This is an important observation because it demonstrates that ER compromises the learning capacity of a neural network.  Specifically, fine-tuning outperforms ER in the last task measure, despite suffering from catastrophic forgetting on past tasks. 
Additionally, it is noteworthy that the gap between the sequential fine-tuning with the neural reasoner of the NeSyBiCL. The two key factors that make the difference are the integration loss and the resetting mechanism, as mentioned in Section \ref{sec:learning}. Our conclusion is that there is a trade-off between addressing catastrophic forgetting and the generalizability of a neural network on a new task. Hence, addressing CL using a neural reasoner may not be the best approach. Our proposed method addresses this problem by leveraging the symbolic knowledge base. Because the burden of CL is handled by the growing knowledge base, the neural reasoner is used to solve the contemporary task. As a result, its learning capacity is not reduced when a large number of tasks are learned. This conclusion is also crucial because it suggests that CL using an end-to-end neural network is not feasible when the tasks become complex compared to the size of the neural reasoner we use.

\subsection{Analytic and Ablative Experiments}\label{sec:analysis}

We present additional experiments to justify why it is necessary to benefit a neuro-symbolic approach for continual learning and to provide a better understanding of the components of our method.

\begin{figure}[t]
\centering
	\begin{subfigure}{.44\textwidth}
		\centering
		\includegraphics[width=0.99\linewidth]{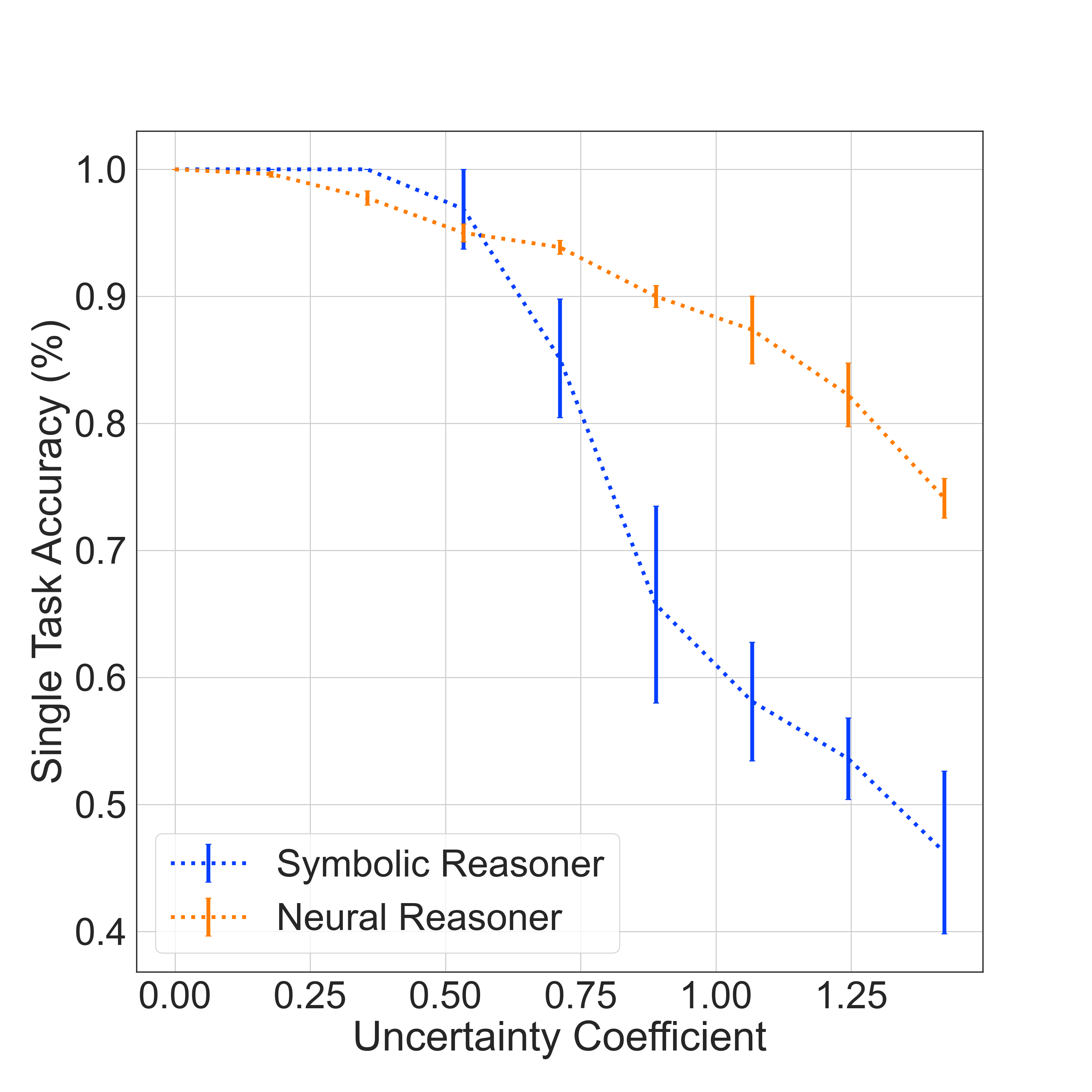}  
		\vspace{-0.35\baselineskip}
	\end{subfigure}
	\begin{subfigure}{.44\textwidth}
		\centering
		\includegraphics[width=0.99\linewidth]{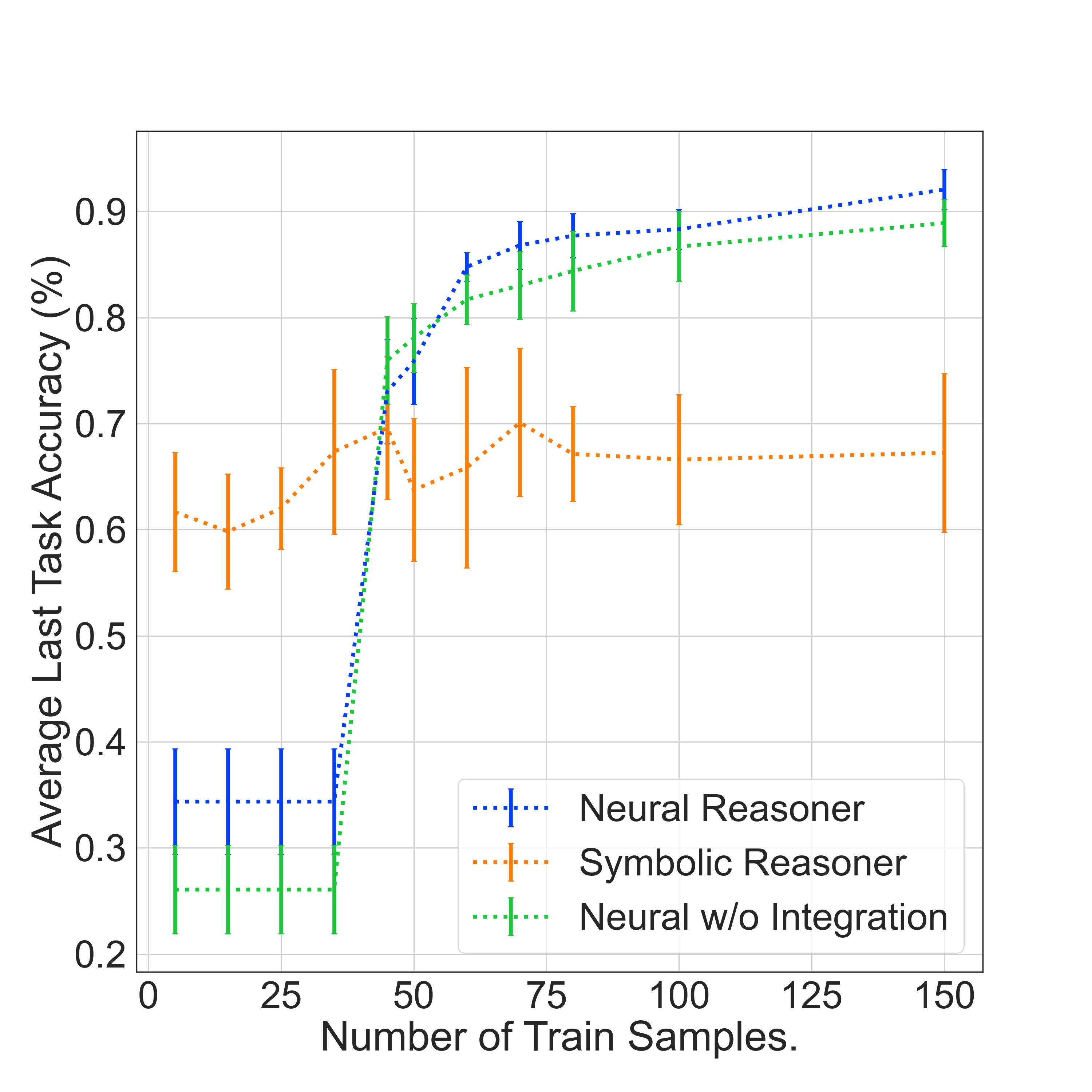}  
		\vspace{-0.35\baselineskip}
	\end{subfigure}
	\caption{
	Ablative experiments: \textbf{(Left)} performance of the neural reasoner and the symbolic reasoner for learning a single task versus uncertainty in the Sketch domain. We see that the neural reasoner is more robust against the injection of uncertainty compared to the symbolic reasoner. \textbf{(Right)} the average accuracy on the last tasks during the continual episode ($\mathcal{A}_last$), versus the number of training samples in the Sketch dataset. We see the effect of the integration loss and also that the performance of the neural reasoner is highly affected by the number of training samples. 
	}\label{fig:unc_and_sample}
    \vspace{-3mm}
\end{figure}

\paragraph{Necessity of neuro-symbolic learning for CL} One natural question is whether a purely symbolic model is sufficient for CL. We observed in Table \ref{tab:result} and Figure \ref{fig:episode}  that NeSyBiC performs slightly better than the symbolic model in the average accuracy measure and outperforms it considerably in the last task accuracy measure. 
To shed light on this behavior, we note that there exists uncertainty and variation in the generated samples of each class. More specifically, there exists a translational noise in the location of each object in the input image. The power of noise controls the variation level in training data. As a result, the abstract graph representation of each image can be more variant, leading to performance degradation in the symbolic reasoner. To study the effect of uncertainty in the input samples, Figure \ref{fig:unc_and_sample}, left,  presents the ability to solve a single task for both models. We observe that the symbolic reasoner is vulnerable with respect to uncertainty, i.e., its generalizability power is less than the neural reasoner.  In contrast, both methods can solve the tasks decently when there exists no uncertainty in the input.
We conclude that a neural reasoner is not going to be the best solution for CL because it cannot perform well to the extent of a neural reasoner in single-task learning. Since neural reasoners are vulnerable with respect to forgetting effects, we conclude that the research focus in CL should be on developing hybrid neuro-symbolic algorithms.

\paragraph{Ablative experiments on the integration Loss} Our motivation for adding the integration loss is to provide a mechanism such that the symbolic reasoner can inject extra supervision into the neural model. This process accords with the interaction of System 1 and System 2 in the nervous system. Figure \ref{fig:unc_and_sample}, right, presents an experiment to demonstrate the effectiveness of the integration loss. In this figure, we have reported the average accuracy for the last tasks versus the number of training samples per class. We first note that the performance of the symbolic reasoner does not considerably depend on the number of training samples. The reason is that it relies on learning the abstract knowledge graph that can be done with a few training points. In contrast, the neural reasoner is highly vulnerable in few-shot learning regimes. The interesting observation is that through using the integration loss,  the performance of the neural reasoner increases significantly.  
This observation suggests that symbolic learning can enable knowledge transfer across tasks because the shared concepts and relations serve as means for transfer learning.

\section{Conclusion}

We developed a brain-inspired continual learning model. Our model is a neuro-symbolic model that consists of a symbolic reasoner,  motivated by System 2 in the brain, and a neural reasoner, to imitate System 1 behavior in the decision-making process of humans. We provide theoretical and empirical studies to demonstrate that the symbolic reasoner mitigates forgetting effects more effectively and the neural reasoner possesses a better generalizability power. The two components complement each other for SOTA performance on two compositional datasets.

An interesting direction for future exploration is extending our framework for solving real-world continual learning problems. Our hypothesis is that our approach will be extendable conditions on using the right symbolic reasoner and benefiting from a carefully annotated dataset for this purpose.

% Our essential hypothesis is that learning a new category of animals or learning a new motory skill happens in the brain by breaking the task into smaller subcomponents and this compositional nature is the key to avoiding forgetting or promoting faster remembering. 

 % \clearpage

 {
\small
\bibliographystyle{plainnat}
\bibliography{main} 
}

\clearpage
\appendix

\section{Proof of the Statement~\ref{statement}}\label{sec:proof}
In this section, we present the justification for Statement~\ref{statement}.
\begin{proof}
 For simplification, we provide the proof in the class-incremental continual learning scenario; However, the proof is easily extendable to the task-incremental or other conditions. More formally, we assume that all the labels for task $t$ comes from the set $C_t$ and these sets are disjoint among different tasks:

$$
C_t = \{y | (x,y) \in \mathcal{D}_t\}, \quad  t \neq t' \Rightarrow C_t \cap C_{t'} = \emptyset
$$

Next, we denote the set of all concatenated training samples as $\mathcal{D} = \bigcup_{t=1,\cdots,T} \mathcal{D}_t$.
To prove that our continual learner has zero forgetting, we have to show that the model that is obtained by the continual training procedure is equivalent to the model obtained in a non-continual setting trained over $\mathcal{D}$. Consider the predictive distributions $p_1(y|x, \text{KB}_1)$ and $p_2(y|x, \text{KB}_2)$ for the two models, respectively. According to   Equation~\eqref{eq:classification}, we can decompose each of the predictive distributions in the following way:

$$
    p(\bm{y} | \bm{x} ; \text{KB}) = \sum_{\mathcal{G} \in \mathbb{G}} p(\bm{y} | \mathcal{G} ; \text{KB}) p(\mathcal{G} | \bm{x}),
$$

According to the persistency assumption, $p(\mathcal{G} | \bm{x})$ is the same for both models. Therefore, we focus on the second term:

$$
p(\bm{y} | \mathcal{G} ; \text{KB}) = \frac{}{}
\text{softmax}(\text{sim}(\mathcal{G},\mathcal{G}_y)) = 
\frac{e^{\text{sim}(\mathcal{G},\mathcal{G}_y)}}{\sum_{(y', \mathcal{G}_{y'}) \in \text{KB}} e^{\text{sim}(\mathcal{G},\mathcal{G}_{y'})}}
\quad \forall (y,\mathcal{G}_y) \in \text{KB}
$$

 where $\text{sim}(\mathcal{G},\mathcal{G}') = \frac{1}{1+d(\mathcal{G},\mathcal{G}')}
$ and $d(\mathcal{G},\mathcal{G}')$ is a distance metric between the two graphs. Thus, it suffices to show that $\text{KB}_1 = \text{KB}_2$ to complete the proof:

$$
\text{KB}_2 = \{(\bm{y}, \mathcal{G}_y) | y \in \mathcal{D}\} = \{(\bm{y}, \mathcal{G}_y) | \bm{y} \in \bigcup_{t=1,\cdots,T} C_t\} 
$$
$$
= \bigcup_{t=1,\cdots,T} \{(\bm{y}, \mathcal{G}_y) | \bm{y} \in C_t\} = \text{KB}_1
$$

In other words, the process of the $\text{KB}$ formation is not affected by the streaming nature of the training data. Therefore, the model experiences no forgetting.
\end{proof}

\section{Integration Loss}\label{sec:int_loss}

To implement the integration loss, first, the color and the shape of each node in the graph are converted to a separate one-hot vector. In other words, each node is described by two one-hot vectors representing its shape and color accordingly. Next, the corresponding vectors of different nodes are added together to construct two vectors. One vector is the summary of the shape of the objects that exist in the image and the other one is representative of the colors. Using the integration loss, the neural reasoner is encouraged to predict the two vectors for each image. It is noteworthy that we set the trade-off hyperparameter between the two loss terms as $\lambda = 1.5$.

\section{Details of the Experimental Setup}\label{sec:exp_details}
In this section, we provide further details on the dataset construction, the symbolic learner, and the neural architectures and hyperparameter values that we used.

\subsection{Dataset}

\begin{figure}[t]
  \centering
  \includegraphics[width=0.99\linewidth]{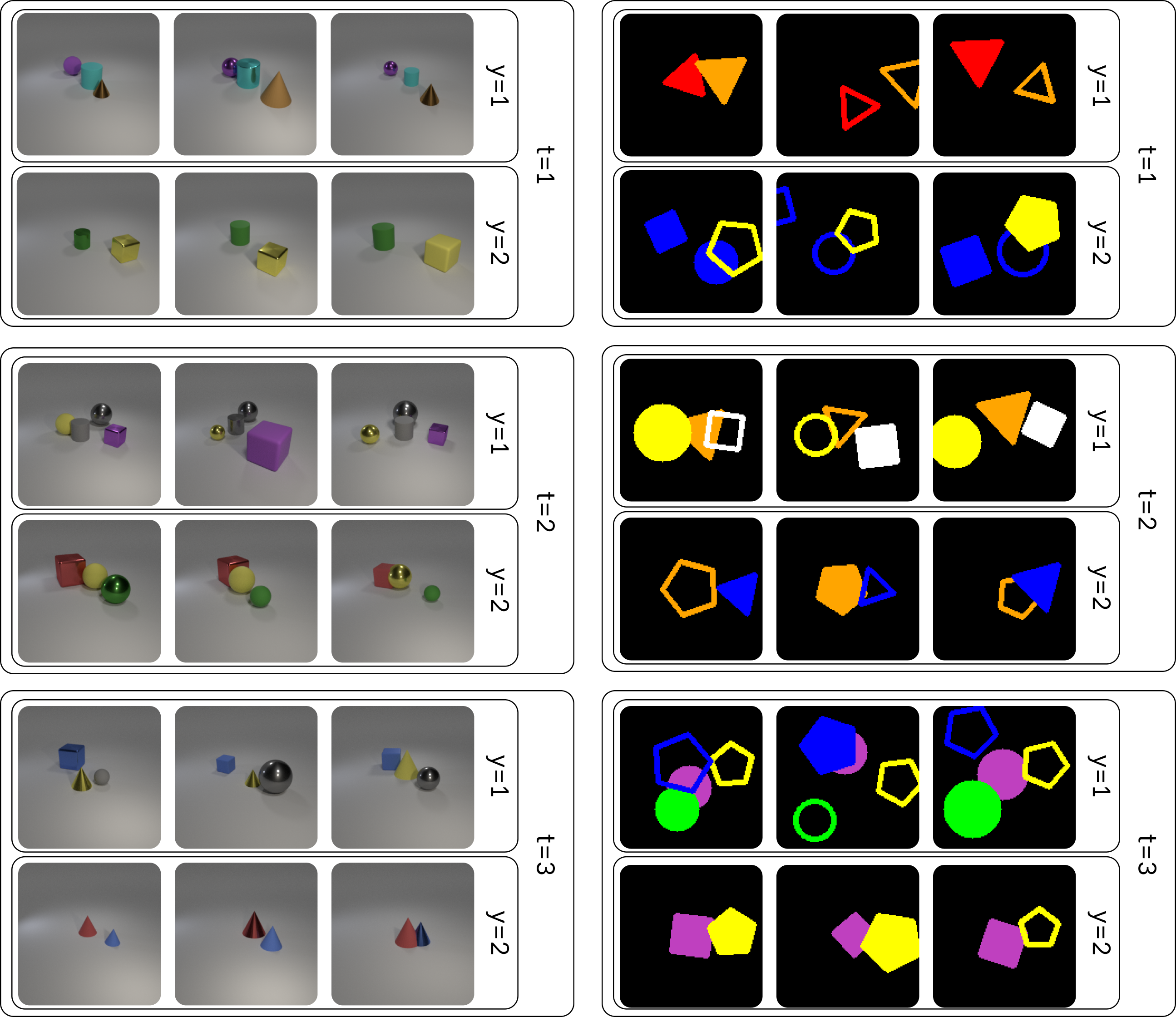}
   \caption{ Sample images from the Clevr \textbf{(Left)} and Sketch \textbf{(Right)} datasets. This figure contains three tasks per dataset and two classes per task.
   }
\label{fig:dataset}  
\end{figure}

 As previously mentioned, each class in this dataset is indicated by the type of the objects, their colors and their bilateral relative locations (Refer to Figure~\ref{fig:dataset} for a visualization).

In the Sketch domain, the shape of objects is uniformly sampled from \textit{\{'Square', 'Triangle', 'Circle', 'Pentagon'\}} and the set of legible colors consists of \textit{\{'Blue', 'Red', 'Purple', 'Green', 'Yellow', 'Orange', 'White'\}}. The objects can be filled with color or not, have different borderline thicknesses, have different area scales, and have different rotation angles. But none of the mentioned factors affect the class label. For the Clevr domain, the objects are selected from the set of \textit{\{'Cone', 'Sphere', 'Cube', 'Cylinder'\}} and the color is selected from 
\textit{\{ 'Gray', 'Red', 'Blue', 'Green', 'Brown', 'Purple', 'Cyan', 'Yellow'\}}. The class-irrelevant factors in this dataset are volume scale, the material of the object (rubber or metal), and rotation angle.

\begin{figure}[t]
  \centering
  \includegraphics[width=0.99\linewidth]{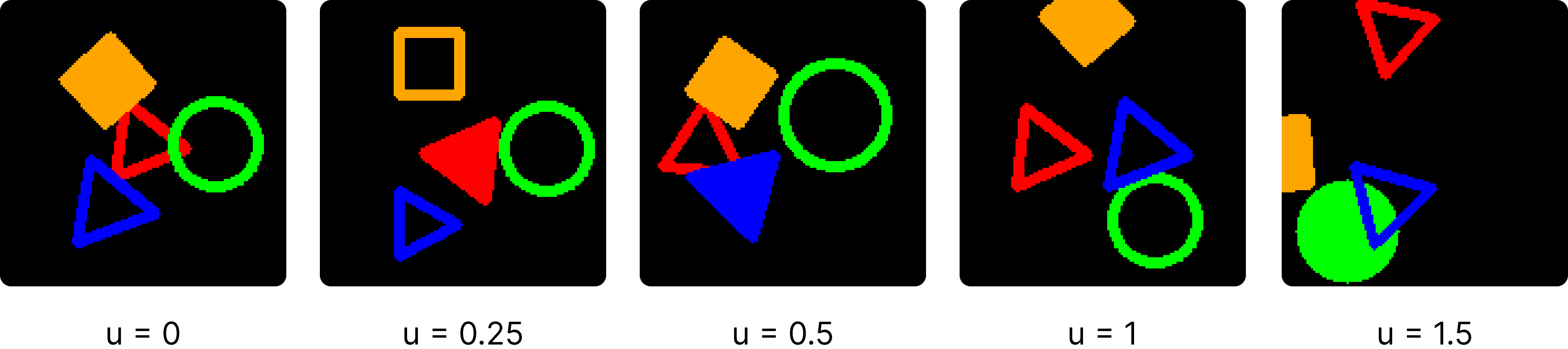}
   \caption{ Sample images with different amounts of translational variation indicated by the scale coefficient $u$.
   }
\label{fig:uncertainty}  
\end{figure}

To create a new class, first, a single object with a certain color is sampled and placed in the center of the scene. Next, one or more objects are selected and distributed on the perimeter of a circle around the first object. Next, to generate inter-class images, the class-irrelevant factors are applied to the image to introduce variations to the samples of the class. Additionally, to inject further uncertainty into the problem, the horizontal and vertical location of each object is modified by a translational noise where the strength of the noise is controlled with a scaling hyperparameter. (See Figure~\ref{fig:uncertainty} for some examples and Section~\ref{sec:analysis} for an analysis).

\subsection{Symbolic Reasoner}

\begin{figure}[t]
  \centering
  \includegraphics[width=0.99\linewidth]{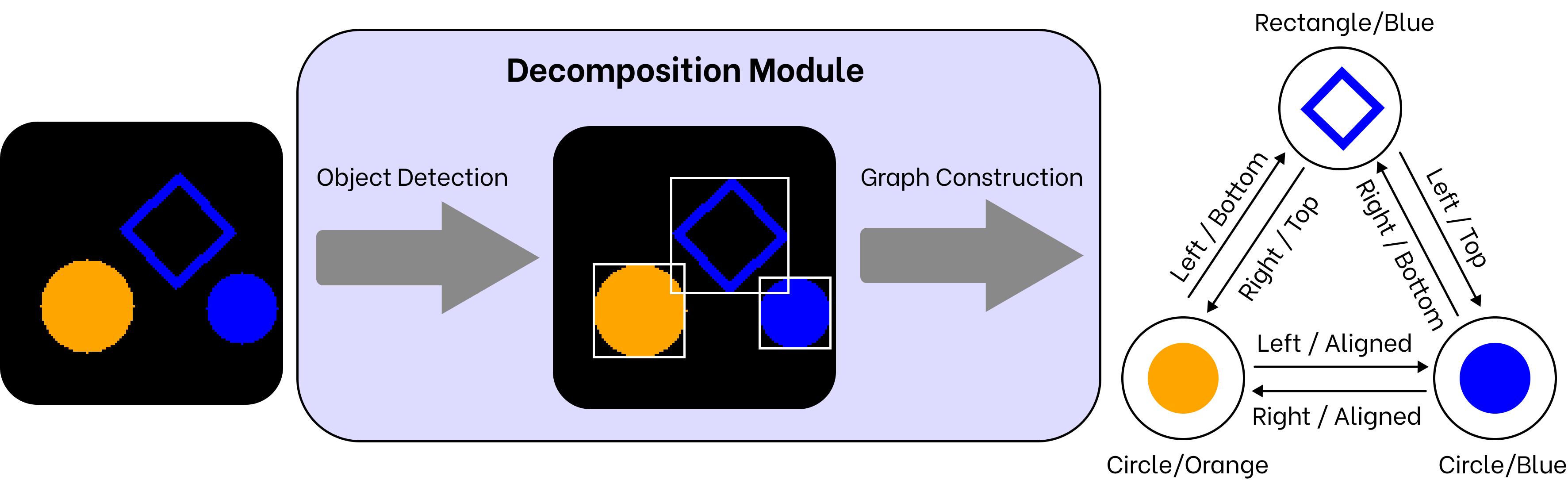}
   \caption{ The procedure of the decomposition module. First, the objects in the image are detected using an object detection algorithm and next, an abstract graph is constructed based on the detections.
   }
\label{fig:symbolic}  
\end{figure}

%In this section, 

We provide details about the symbolic learner and its sub-elements, i.e., the decomposition module and the graph construction process. See Figure~\ref{fig:symbolic} for an overview of the decomposition module. This module first decomposes the input image using an object detection algorithm, e.g. a Faster-RCNN \cite{fasterRCNN}. In addition to the bounding box, the object detector can effectively identify the color and shape of each object using a classification head on top of the object detector. Using the extracted information, the corresponding graph is constructed where each node represents the objects (subconcepts) in the image and edges represent the location of each concept in the scene when relatively compared to other nodes. As mentioned in the main text, after decomposing images in the dataset, further processing steps are taken to compute the distance between different graphs of the input images during the learning and inference phases. It is noteworthy that we used graph edit distance as a metric to compute the distance between two graphs in the symbolic reasoner.

The Faster-RCNN used in this work has two layers of CNN with the kernel size of 5 as the backbone, followed by RPN, ROI pooling, and classification layers. Table~\ref{tab:fasterrcnn} contains the results of training this module over the pretraining set. Moreover, the inference time of the algorithm is reported in the same table. The high processing time of this module is a considerable practical barrier to using the symbolic reasoner. Therefore, one might preprocess the whole training and testing corpus priorly to minimize the object detection time overhead. However, due to the high performance of the module on both datasets and because of the fact that the target of this research is fundamentally different from developing accurate object detectors, one might assume that the symbolic reasoner has oracle access to the output of an approximately perfect object detector.

\begin{table}[!t] 
  \centering
\begin{tabular}{cccc}
\specialrule{1pt}{2.1pt}{2.5pt}
\phantom{Dataset} & Average Precision (\%) & Average Recall (\%) & Inference Time \\
Dataset & @ IOU = 0.5 & @ IOU = 0.5 & per 1000 Samples (s)  \\
\cmidrule(r){1-4} 
Clevr                      & $93.7$           & $97.9$            & $22.0$\\
Sketch         & $97.8$           & $99.1$            & $21.6$ \\ \bottomrule
\end{tabular} \vspace{0.9em}
\caption{The results of training Faster-RCNN on the pertaining set. The inference time is reported on an NVIDIA V100 GPU.}
  \label{tab:fasterrcnn}
\end{table}

\subsection{Architecture and Hyperparameters}

In the experiments, we used two datasets in the Sketch and Clevr domains. The images from the two datasets have a resolution of $64 \times 64$ and $160 \times 120$ respectively. The neural learners used in this work are constructed from 2 layers of convolutional neural networks with a kernel size of 5 as the preliminary feature extractors. On top of that, two MLP layers are placed where the first one is responsible for encoding the input in the embedding space and the second one is considered as the classifier weights. We optimized neural architectures using the Adam optimizer and a learning rate of $10^{-3}$ for 200 epochs per task. It is noteworthy that each experiment is repeated 4 times and its average and variance is reported. 

In most of the experiments, we used a training dataset with 10 tasks and 10 classes per task. We used 200 train and 50 test samples per class. Additionally, we have also used a pretraining set of 50 classes to train the Faster-RCNN module or to pretrain the feature extractor backbones.

\section{Additional Analytic Experiments}

In this section, we offer additional experiments that provide a deeper insight about our algorithm.

\subsection{Running Times Comparison}

We compare the training and testing time of the neural and symbolic reasoner to demonstrate a crucial advantage of the neural reasoner compared to the symbolic reasoner which is its inference time. Table~\ref{tab:runtime} provides a comparison on the Sketch dataset. Based on this table, the symbolic reasoner has much slower training and inference procedures. One possible explanation for such a difference could be the distinct nature of the two algorithms. Neural networks benefit from highly-parallelized data processing pathways implemented on GPU infrastructures. However, the symbolic algorithm has a sequential nature, especially in calculating the graph distances using the edit distance metric. This observation also aligns with similar behavior in humans, where it is reported that System 1 is much faster in decision-making compared to System 2 \cite{kahneman2011thinking}. 

\begin{table}[!t] 
  \centering
\begin{tabular}{ccc}
\specialrule{1pt}{2.1pt}{2.5pt}
Model & Training Time (s) & Inference Time (s) \\
\cmidrule(r){1-3} 
Neural Reasoner                      & $13.2$           & $0.04$ \\
Symbolic Reasoner         & $135.7$           & $41.5$            \\ \bottomrule
\end{tabular} \vspace{0.9em}
\caption{Running time of implemented models measured on an NVIDIA V100 GPU. Both models were trained on a single-task Sketch dataset containing 10 classes and 100 train samples per class. The running time of the neural reasoner is measured for 200 epochs. The testing set for both models also includes 100 samples per class. The time measured for the symbolic reasoner includes the Faster-RCNN inference in addition to the symbolic reasoning performed on the extracted graphs.}
  \label{tab:runtime}
\end{table}

\subsection{Long Continual Episode on Clevr Dataset}

\begin{figure}[t]
	\begin{subfigure}{.48\textwidth}
		\centering
		\includegraphics[width=0.99\linewidth]{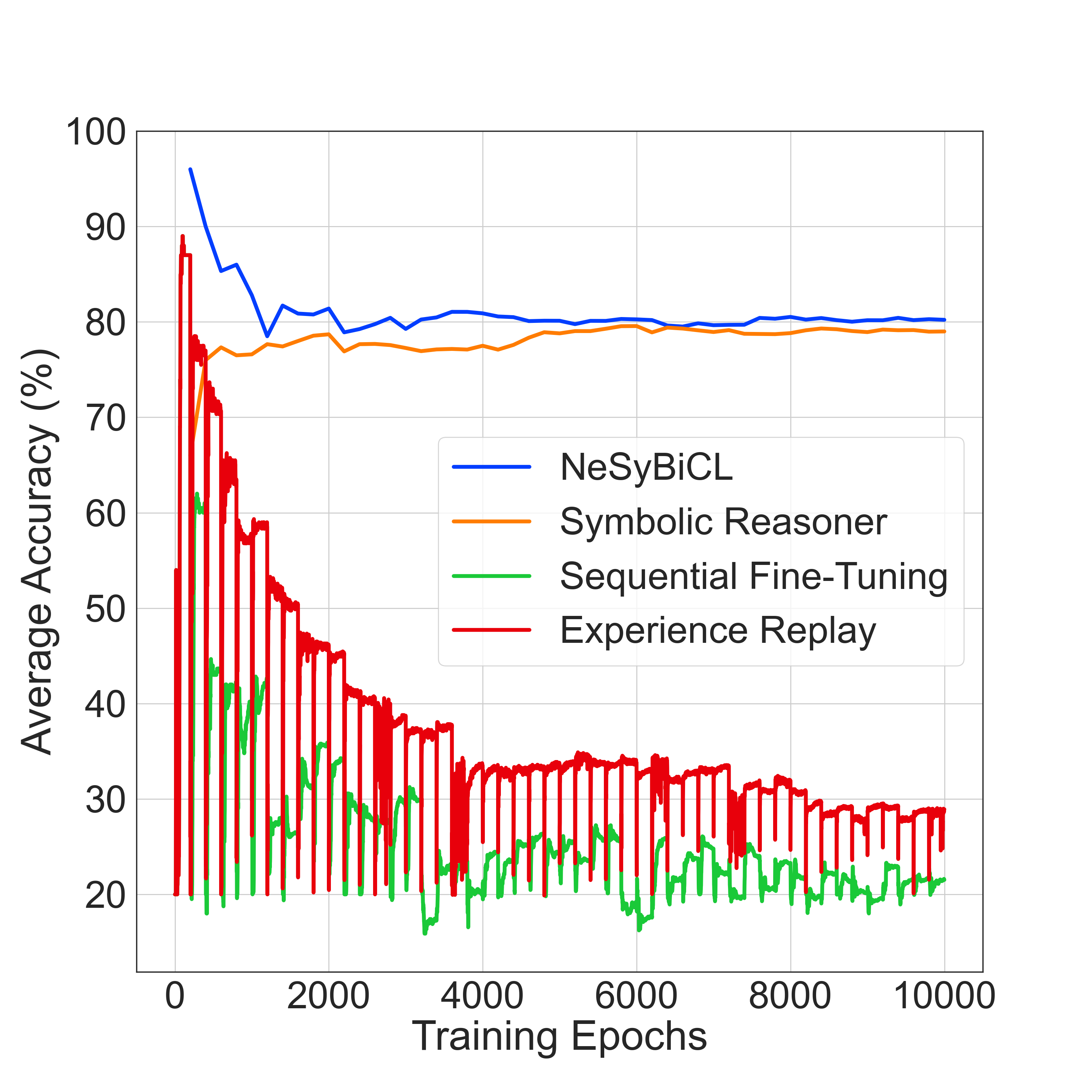}  
		\vspace{-0.35\baselineskip}
	\end{subfigure}
	\begin{subfigure}{.48\textwidth}
		\centering
		\includegraphics[width=0.99\linewidth]{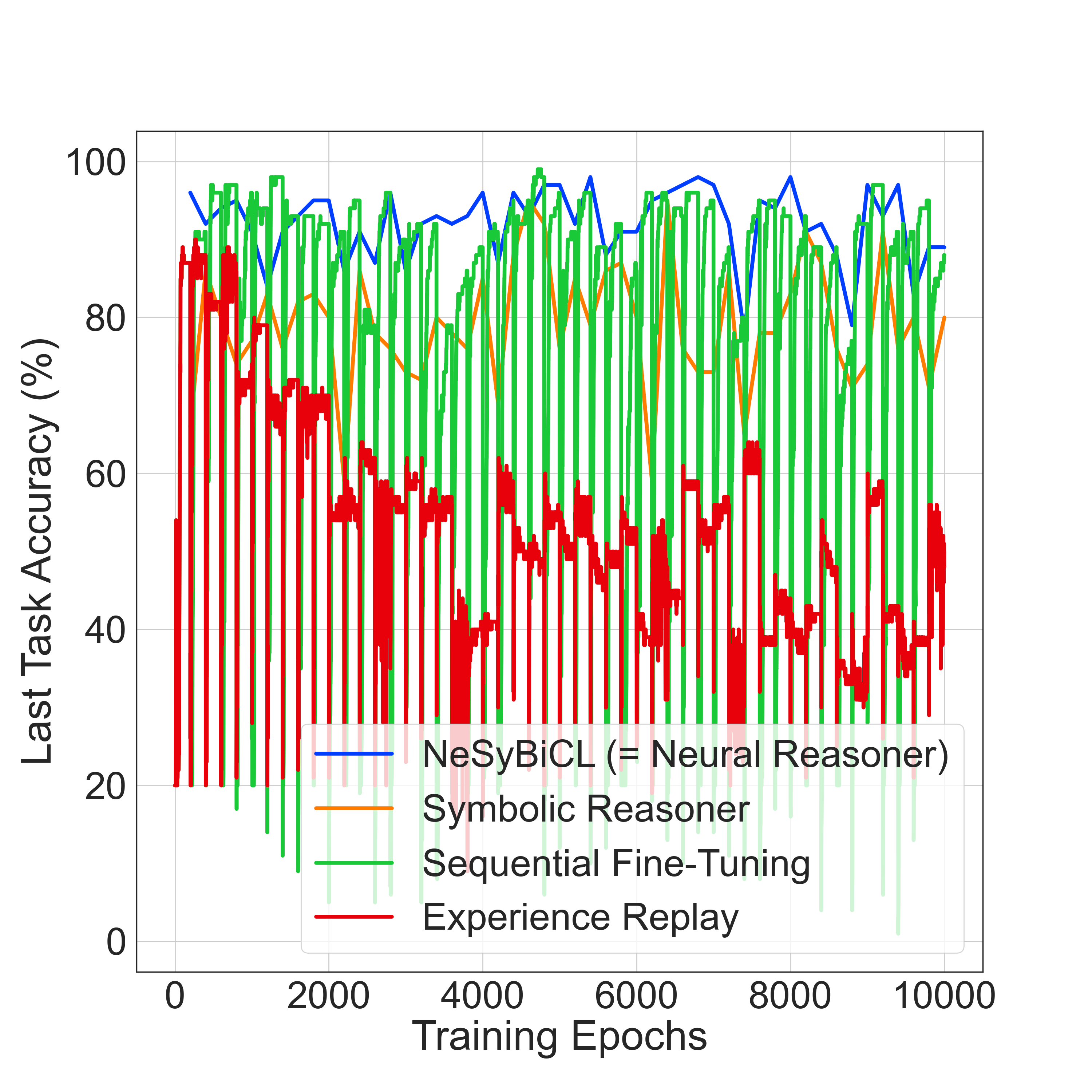}  
		\vspace{-0.35\baselineskip}
	\end{subfigure}
	\caption{
	Learning curves for a long continual learning episode consisting of 50 tasks in the Clevr domain:  \textbf{(Left)} the average accuracy on all previous tasks at each time step. We observe that NeSyBiC and the symbolic reasoner lead to minimal forgetting effects. \textbf{(Right)} the accuracy on the test samples of the last task at each timestep. We observe that NeSyBiC achieves better performance when solving the contemporary task compared to the symbolic learner. The only difference is that in this experiment, each task contains 5 classes as compared to the 10 classes in Figure~\ref{fig:episode}.
	}
	\label{fig:clevrlongrun}
\end{figure}

 Figure~\ref{fig:episode} contains learning curves for a long continual episode containing 50 tasks of the Sketch dataset. Here, we repeat a similar experiment on the Clevr domain to further emphasize the effectiveness of our model in mitigating catastrophic forgetting. It can be similarly observed in Figure~\ref{fig:clevrlongrun} that both the hybrid and symbolic learners are not affected by catastrophic forgetting. In contrast, the neural-based models greatly suffer from the forgetting effect. In addition to that, The NeSyBiCL model performs considerably better than the symbolic learner in the last task accuracy measure.

\subsection{Effect of Pretraining}

We investigate the effect of using a pretrained backbone with other CL baselines. To that end, 
a convolutional neural network is trained on the pretraining corpus using the typical cross-entropy loss. Next, the convolutional layers are frozen and used as the initial point of the backbone of the neural learner in a CL episode. Table~\ref{tab:pretrain} demonstrates the results of this procedure. The table completely follows the setup of Table~\ref{tab:result}, except for the fact that the backbone is pretrained. It is noticeable that using such a technique could not greatly affect most of the baselines.

\begin{table}[!t] 
  \centering
\begin{tabular}{@{}ccccc@{}}
\specialrule{1pt}{2.1pt}{2.5pt}
\phantom{Methods} & \multicolumn{2}{c}{Sketch} & \multicolumn{2}{c}{Clevr}        \\ \cmidrule(r){2-3} \cmidrule(r){4-5} 
Methods                  & $\mathcal{A}_{all}$(\%)   & $\mathcal{A}_{last}$(\%)   & $\mathcal{A}_{all}$(\%)   & $\mathcal{A}_{last}$(\%) \\ \midrule
Fine-Tuning                                  & $17.7 \pm 1.5$           & $85.8 \pm 8.6$            & $21.0 \pm 0.1$              & $89.6 \pm 5.0$               \\
Multi-Task                                   & $90.7 \pm 0.2$           & $-$                       & $91.5 \pm 0.1$              & $-$                          \\\cmidrule(r){1-5} 
EWC \cite{kirkpatrick2017overcoming}         & $20.5 \pm 0.8$           & $88.9 \pm 5.2$            & $22.9 \pm 0.2$              & $88.7 \pm 5.5$               \\
ER \cite{rolnick2019experience}              & $29.8 \pm 2.1$           & $80.8 \pm 8.3$            & $30.3 \pm 0.4$              & $82.4 \pm 7.3$               \\
SI \cite{huttenlocher1979synaptic}           & $19.7 \pm 0.1$           & $87.5 \pm 4.3$            & $19.8 \pm 0.2$              & $87.1 \pm 4.4$               \\
LWF \cite{LiH16eLWF}                         & $19.1 \pm 1.7$           & $87.1 \pm 6.4$            & $21.1 \pm 0.1$              & $87.9 \pm 6.6$               \\
GEM \cite{lopez2017gradient}                 & $18.6 \pm 1.0$           & $87.9 \pm 5.6$            & $21.8 \pm 0.1$              & $89.5 \pm 5.1$               \\ 
\bottomrule
\end{tabular} \vspace{0.9em}
\caption{Performance comparison of different baselines on the two compositional datasets when a pretrained backbone is used.  }
  \label{tab:pretrain}
  \vspace{-0.9em}
\end{table}

\section{Infrastructures and Data Availability}
% Related source code is submitted accordingly. In this repository, there exists a README file containing instructions and configuration details. Moreover, the licenses of the freely available datasets and used source codes are also available in the README file. Most of the code in this repository is implemented using the PyTorch framework. 
The Sketch dataset is generated using Open-CV and the Clevr dataset is generated using the Blender toolkit. Moreover, we used an NVIDIA P100 GPU and a Google Colab service with an NVIDIA V100 GPU. As a rough estimation, all of the experiments of Table~\ref{tab:result} can be replicated in approximately 6 hours on a V100 GPU.

\end{document}